\documentclass{article}

\usepackage[final]{corl_2026}

\title{ExecGap: Exploiting Command-Execution Gap for Patch Attacks on Vision-Language-Action Models}

\author{
  Keke Tang$^{1,2*\dagger}$, \ \ \ Guolin Qiu$^{1,2*}$, \ \ \ Tianyu Hao$^{1}$ \\
  \textbf{Daizong Liu$^{3}$, \ \ \ Yikai Zhang$^{1\dagger}$,  \ \ \ Peican Zhu$^{4}$,  \ \ \ Zhihong Tian$^{1}$}\\
  $^{1}$Guangzhou University  \ \ \ 
  $^{2}$4Axiom Robotics\\ $^{3}$Wuhan University \ \ \ 
  $^{4}$Northwestern Polytechnical University\\ 
  $^*$Joint first authors  \ \  \
  $^\dagger$Corresponding authors\\
  \\
  {\small\normalfont\texttt{tangbohutbh@gmail.com, 2112533042@e.gzhu.edu.cn}}\\
  {\small\normalfont\texttt{howty666@gmail.com, daizongliu@whu.edu.cn}}\\
  {\small\normalfont\texttt{ykzhang@gzhu.edu.cn, ericcan@nwpu.edu.cn, tianzhihong@gzhu.edu.cn}}
}

\usepackage{algorithm}
\usepackage{algpseudocode}
\usepackage{xcolor}
\usepackage{amsmath}
\usepackage{amssymb}
\usepackage{censor}
\usepackage{bm}
\usepackage{mathtools}
\usepackage{enumitem}

\usepackage{booktabs}      
\usepackage{multirow}      
\usepackage{graphicx}      
\usepackage{dblfloatfix}

\usepackage{subcaption}
\usepackage{siunitx}  

\usepackage{caption}  

\providecommand{\revise}[1]{{#1}}
\newcommand{\firstpara}[1]{\noindent\textbf{{#1}.}~}

\begin{document}

\maketitle


\begin{abstract}
Vision-Language-Action (VLA) models have become a powerful paradigm for robotic manipulation and show strong promise for real-world deployment, making adversarial robustness critical for reliable robot operation.
Existing adversarial attacks on VLA-controlled robots mainly optimize prediction-level discrepancies or rollout-level deviations to disrupt manipulation.
However, they rarely account for the fact that VLA-predicted commands are not executed directly, but must be realized through downstream control and robot embodiment.
In this paper, we propose ExecGap, an adversarial patch attack framework that exploits this command-execution gap to expose embodiment-specific execution vulnerabilities.
Specifically, ExecGap biases VLA-predicted task-space commands toward low-executability directions, which are estimated from kinematic inefficiency and dynamic weak actuation.
This induces execution degradation, where the predicted command becomes difficult to realize effectively, thereby impairing task completion.
Extensive simulation experiments across multiple VLA policies show that ExecGap induces larger command-execution mismatch and weaker realized execution than existing patch attacks.
Real-robot experiments further validate that the deployed patch can induce execution degradation on physical platforms.

\end{abstract}



\section{Introduction}

Vision-Language-Action (VLA) models have emerged as a powerful paradigm for robotic manipulation~\cite{kim2025openvla, zitkovich2023rt, black2024pi_0}. 
By integrating visual perception, language conditioning, and action generation, they enable robots to interpret natural-language instructions and perform diverse manipulation tasks. 
As VLA-based robots show growing promise for real-world deployment, their reliability under adversarial conditions becomes a critical requirement for safe operation in physical environments.

Recent studies have begun to examine this concern through adversarial attacks on robotic VLA systems. 
Existing attacks mainly optimize visual perturbations or adversarial patches to induce prediction-level discrepancies, semantic confusion, or rollout-level deviations~\cite{wang2025exploring, sadeghi2023attacking, wu2024characterizing}. 
Typical objectives enlarge the difference between clean and perturbed actions, drive the policy toward erroneous manipulation behaviors, or cause visible deviations over the rollout. 
Although effective at task disruption, these methods primarily treat the VLA output or rollout behavior as the attack target, while the physical realization of the predicted command remains less explored.

\begin{figure}[!t]
    \centering
    \includegraphics[width=0.9\linewidth]{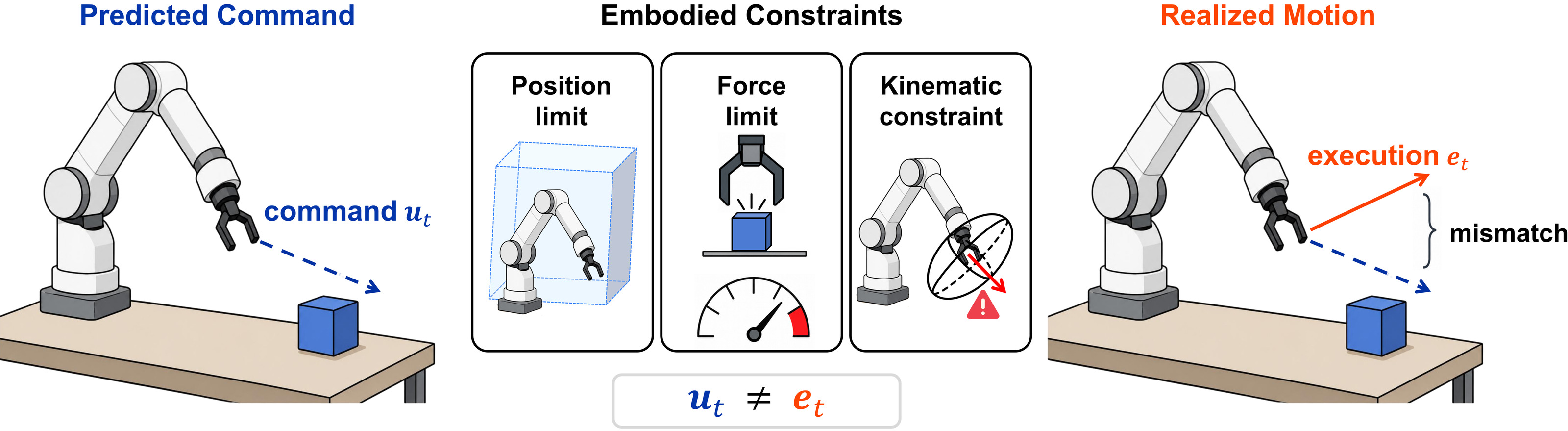}
\caption{
Command-execution gap in VLA-controlled manipulation.
A VLA model predicts a task-space command $\mathbf{u}_t$ from visual observation and language instruction.
During embodied execution, constraints from control, kinematics, and actuation can reshape or attenuate the command, causing command-execution mismatch between the task-space command and the realized motion.
}
    \label{fig:concept}
\end{figure}

However, VLA-controlled manipulation is not determined by prediction alone. 
A VLA model may output a semantically plausible command, yet this command must still be realized through downstream control and robot embodiment. 
As illustrated in Fig.~\ref{fig:concept}, embodied constraints such as position limits, force limits, and kinematic constraints can reshape or attenuate the task-space command before physical execution. 
This creates a command-execution gap: a command that appears reasonable at the prediction level may still be difficult to realize effectively, thereby reducing effective task progress and making task completion less reliable.


\revise{In this paper, we expose an embodiment-specific attack surface arising from the gap between VLA-issued commands and their physical realization, and study the resulting failure mode termed execution degradation.}
Rather than directly maximizing prediction discrepancy or rollout-level deviation, our goal is to enlarge the mismatch between the VLA-predicted task-space command and the realized motion.
To this end, we propose ExecGap, an adversarial patch attack framework that exploits the limited executability of robot motion. 
Specifically, ExecGap derives low-executability directions from two factors: kinematic inefficiency and dynamic weak actuation. 
By biasing commands toward these hard-to-execute directions, ExecGap increases command-execution mismatch, weakens realized task progress, and degrades task completion. 
Extensive simulation experiments on LIBERO suites~\cite{liu2023libero} show that ExecGap effectively disrupts manipulation and induces larger command-execution mismatch with weaker realized execution across OpenVLA~\cite{kim2025openvla}, OpenVLA-OFT~\cite{kim2025fine}, and $\pi_0$~\cite{black2024pi_0}.
Real-robot validation further shows that physically deployed patches can induce execution degradation on robotic hardware.

Overall, our contributions are three-fold:
\begin{itemize}[topsep=0pt, itemsep=1pt, parsep=0pt, partopsep=0pt] 
\item We identify execution degradation as an execution-side VLA failure mode arising from degraded embodied realization rather than  prediction-side discrepancy alone.
\item We propose ExecGap, a robot-aware patch attack that exploits kinematic inefficiency and dynamic weak actuation to induce low-executability commands.
\item We validate ExecGap in simulation and on real robots, showing stronger execution degradation in simulation and physical feasibility on real platforms.
\end{itemize}

\section{Related Work}

\firstpara{Vision-Language-Action Models}
VLA models integrate visual perception, language understanding, and action generation for robotic manipulation.
Representative systems, including RT-2~\cite{zitkovich2023rt}, OpenVLA~\cite{kim2025openvla}, Octo~\cite{team2024octo}, and $\pi_0$~\cite{black2024pi_0}, adapt large-scale pretrained representations to robot control and enable language-conditioned manipulation across diverse tasks~\cite{o2024open,liu2023libero,mees2022calvin}.
Broader discussions of recent VLA models and deployment challenges can be found in surveys~\cite{ma2026survey,kawaharazuka2025vision,wang2025vlatest}.


\firstpara{Adversarial attacks on robotic  systems}
\revise{Adversarial attacks have been studied in visual perception~\cite{szegedy2013intriguing,tang2022rethinking,Tang-2023-ManifoldAttack,Tang-2024-FLAT,tang2026less} and robotic manipulation~\cite{wang2025advgrasp,tang2026occluding}, with growing attention to VLA-based robots.}
Recent studies show that VLA systems are vulnerable to visual perturbations, adversarial patches, object-level modifications, language prompts, and multimodal attacks~\cite{wang2025exploring,xu2025model,lu2025robots,chen2026tex3d,jones2025adversarial,yan2025alignment,wang2026partially}.
Existing attacks induce erroneous actions, manipulation failures, or rollout-level deviations by corrupting observations or instructions.
In contrast, ExecGap targets execution degradation by steering VLA-predicted commands toward low-executability directions, while considering how commands are physically realized.

\firstpara{Execution difficulty in embodied robot control}
Execution difficulty has long been studied in robot control through manipulability, force capability, and dynamic constraints~\cite{siciliano2009robotics,lynch2017modern,yoshikawa1985manipulability,yoshikawa1985dynamic,chardonnet2009force,asada1983geometrical}.
These analyses have been used for motion planning, redundancy resolution, singularity avoidance, grasping, contact-rich manipulation, and whole-body control~\cite{nakamura1987task,rozo2017learning,chan1995weighted,chiaverini1997singularity,gravagne2001manipulability,khatib2004whole,sentis2005synthesis,khatib1987unified}.
While prior work uses executability analysis to improve feasibility and control performance, we use it adversarially to expose execution-side vulnerabilities in VLA-controlled robots.

\section{Problem Formulation}
\label{sec:problem}

\subsection{VLA Policy and Embodied Execution}

\firstpara{VLA policy}
We consider a vision-language-action (VLA) policy $\Phi$.
At step $t$, given a visual observation $I_t$, a language instruction $L$, and an optional robot state $s_t$, the policy outputs a model-specific command.
Since VLA models may use different action parameterizations, such as continuous actions, tokenized actions, or flow-based actions, we compare commands and realized motions in a common task space:
\begin{equation}
\mathbf{c}_t=\Phi(I_t,L,s_t)\in\mathcal{C},
\qquad
\mathbf{u}_t=\Pi(\mathbf{c}_t,s_t)\in\mathbb{R}^{d},
\end{equation}
\revise{
where $s_t$ denotes the robot state and is omitted from the policy input when not required,}
$\mathcal{C}$ is the model-specific action space, and $\Pi$ maps $\mathbf{c}_t$ to a comparable task-space command, such as translational end-effector motion.

\firstpara{Embodied execution}
The downstream controller and robot embodiment realize the command as a task-space motion over one control interval:
\begin{equation}
\mathbf{e}_t=\mathcal{E}(\mathbf{c}_t,s_t)\in\mathbb{R}^{d},
\end{equation}
where $\mathcal{E}$ abstracts control, kinematics, dynamics, and physical constraints during execution.
In general, $\mathbf{e}_t\neq\mathbf{u}_t$, since embodiment constraints may reshape, attenuate, or redirect the commanded motion before it is physically realized.

\subsection{Adversarial Patch Attack}

\firstpara{Attack setting}
Following the adversarial patch threat model~\cite{brown2017adversarial}, we apply a static visual patch $P\in\mathcal{P}$ to the VLA observation.
Given a patch location $(x,y)$, the adversarial observation, command, and task-space command are
\begin{equation}
I_t^{\mathrm{adv}}(P)=\mathcal{A}(I_t,P,x,y),
\quad
\mathbf{c}_t^{\mathrm{adv}}(P)=\Phi(I_t^{\mathrm{adv}}(P),L,s_t),
\quad
\mathbf{u}_t^{\mathrm{adv}}(P)=\Pi(\mathbf{c}_t^{\mathrm{adv}}(P),s_t).
\end{equation}
We consider a robot-aware white-box patch attack.
\revise{The attacker has white-box access to the VLA model, robot-state information, robot kinematic and dynamic parameters, and clean expert trajectories for offline patch optimization, but can only modify the visual observation through the patch.}
The instruction, VLA parameters, downstream controller, robot state, and robot dynamics are not modified by the attacker.

\firstpara{Standard objective}
Conventional patch attacks optimize prediction-level or task-level objectives,
\begin{equation}
P_{\mathrm{atk}}^*
=
\arg\max_{P\in\mathcal{P}}
\mathcal{L}_{\mathrm{atk}}(P),
\end{equation}
where $\mathcal{L}_{\mathrm{atk}}$ typically measures action discrepancy, targeted deviation, or task failure.
Such objectives attack the prediction side of the VLA policy but do not explicitly model whether the predicted command can be effectively realized by the robot embodiment.

\subsection{Execution Degradation Objective}

\firstpara{Command-execution mismatch}
Instead of directly optimizing rollout-level deviation, we formulate execution degradation through command-execution mismatch.
Under attack, the realized task-space motion and the command-execution mismatch are
\begin{equation}
\mathbf{e}_t^{\mathrm{adv}}(P)
=
\mathcal{E}(\mathbf{c}_t^{\mathrm{adv}}(P),s_t),
\qquad
\mathcal{L}_{\mathrm{mis}}(t;P)
=
\left\|
\mathbf{u}_t^{\mathrm{adv}}(P)
-
\mathbf{e}_t^{\mathrm{adv}}(P)
\right\|_2 .
\end{equation}
A persistent mismatch indicates that the VLA command is not effectively realized by the robot, 
which can reduce effective task progress even when the command is valid in task space.


\firstpara{Attack objective}
The execution degradation objective seeks
\begin{equation}
P^*
=
\arg\max_{P\in\mathcal{P}}
\mathbb{E}_t
\left[
\mathcal{L}_{\mathrm{mis}}(t;P)
\right].
\end{equation}
This objective captures an execution-side failure mode by enlarging the mismatch between the task-space command and the realized motion.

\section{Method}
\label{sec:method}

\begin{figure}[!t]
    \centering
    \includegraphics[width=1\linewidth]{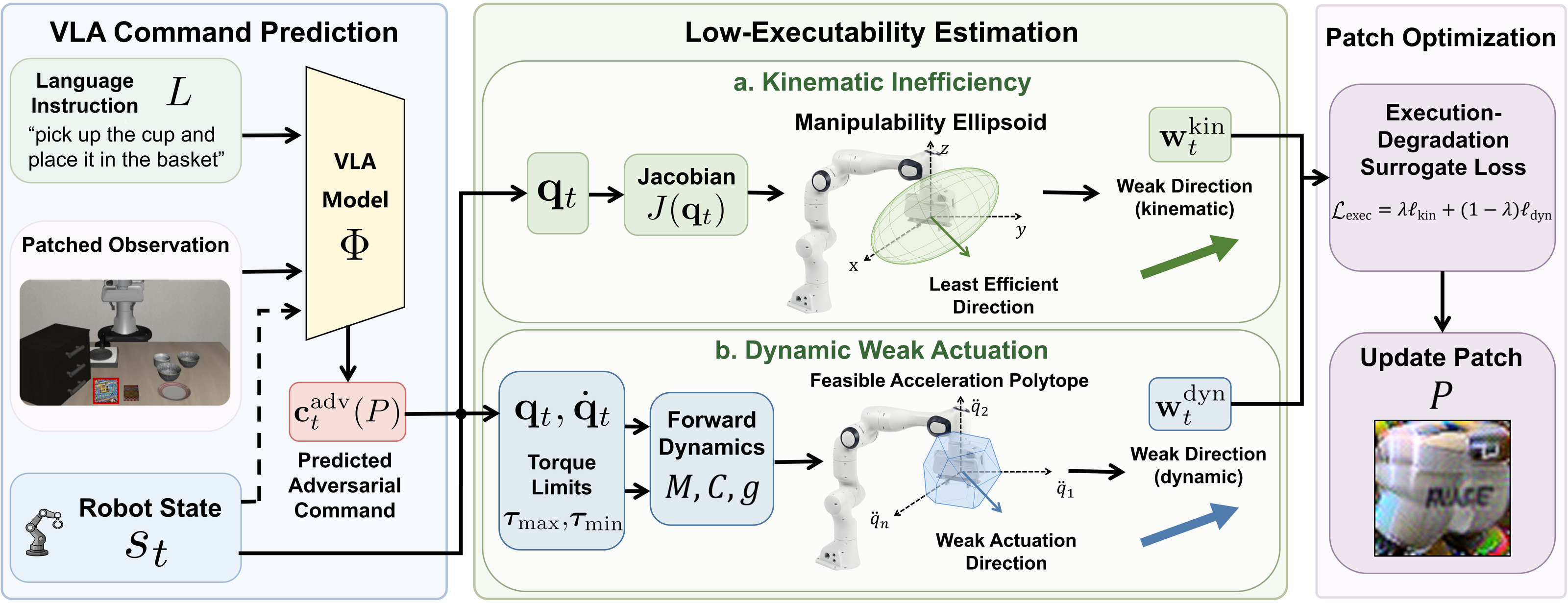}
\caption{
Overview of ExecGap.
ExecGap first estimates low-executability directions from kinematic inefficiency and dynamic weak actuation.
It then optimizes a visual patch to bias the VLA-predicted task-space command toward these directions, enlarging command-execution mismatch and inducing execution degradation.
}
    \label{fig:overview}
\end{figure}


In this section, we present ExecGap, a patch attack that exploits the command-execution gap to induce execution degradation in VLA-controlled robots, as shown in Fig.~\ref{fig:overview}.

Directly optimizing the mismatch objective in Sec.~\ref{sec:problem} is challenging because it requires differentiating through downstream execution and closed-loop rollout dynamics.
ExecGap therefore uses a robot-aware surrogate: it estimates low-executability task-space directions from kinematic inefficiency (Sec.~\ref{sec:ki}) and dynamic weak actuation (Sec.~\ref{sec:dwa}), then optimizes the adversarial patch to bias VLA-predicted commands toward these weakly realized directions (Sec.~\ref{sec:po}).
This enlarges command-execution mismatch without differentiating through execution.

\subsection{Low-Executability via Kinematic Inefficiency}
\label{sec:ki}

\firstpara{Kinematic inefficiency}
Let $\mathbf{q}_t$ denote the robot joint configuration at the current step.
Following standard differential kinematics~\cite{yoshikawa1985manipulability,spong2020robot}, we use
\begin{equation}
    \dot{\mathbf{x}}
    =
    J_t\dot{\mathbf{q}},
    \qquad
    J_t
    =
    J(\mathbf{q}_t)
    \in
    \mathbb{R}^{d\times n},
\end{equation}
where $\dot{\mathbf{x}}\in\mathbb{R}^{d}$ is the task-space velocity, $\dot{\mathbf{q}}\in\mathbb{R}^{n}$ is the joint velocity, and $n$ is the number of joints.
The directional capability of $J_t$ is characterized by singular value decomposition:
\begin{equation}
    J_t
    =
    U_t\Sigma_tV_t^{\top}.
\end{equation}
Let $\sigma_{t,i}$ and $\mathbf{u}_{t,i}$ denote the $i$-th singular value and the corresponding left singular vector.
The kinematically inefficient direction is defined as
\begin{equation}
    \mathbf{w}_t^{\mathrm{kin}}
    =
    \mathbf{u}_{t,i^*},
    \qquad
    i^*
    =
    \arg\min_i \sigma_{t,i}.
\end{equation}
Motion along $\mathbf{w}_t^{\mathrm{kin}}$ requires relatively large joint-space motion while producing limited task-space progress.

\firstpara{Kinematic degradation loss}
Given the adversarial task-space command $\mathbf{u}_t^{\mathrm{adv}}(P)$, we encourage it to align with the kinematically inefficient direction:
\begin{equation}
    \ell_{\mathrm{kin}}
    \left(
        \mathbf{u}_t^{\mathrm{adv}}(P),
        \mathbf{w}_t^{\mathrm{kin}}
    \right)
    =
    -
    \frac{
    \left|
    \left(\mathbf{u}_t^{\mathrm{adv}}(P)\right)^{\top}
    \mathbf{w}_t^{\mathrm{kin}}
    \right|
    }
    {
    \left\|\mathbf{u}_t^{\mathrm{adv}}(P)\right\|_2
    \left\|\mathbf{w}_t^{\mathrm{kin}}\right\|_2
    +
    \epsilon
    },
\end{equation}
where $\epsilon$ is a small constant.
The absolute value treats opposite directions along $\mathbf{w}_t^{\mathrm{kin}}$ equivalently, since both correspond to low kinematic capability.

\subsection{Low-Executability via Dynamic Weak Actuation}
\label{sec:dwa}

\firstpara{Dynamic weak actuation}
Execution difficulty also depends on the robot's dynamic actuation capability.
For the robot state $(\mathbf{q}_t,\dot{\mathbf{q}}_t)$, we consider the standard rigid-body dynamics~\cite{spong2020robot,featherstone2008rigid}:
\begin{equation}
    M(\mathbf{q}_t)\ddot{\mathbf{q}}
    +
    \mathbf{h}(\mathbf{q}_t,\dot{\mathbf{q}}_t)
    =
    \boldsymbol{\tau},
    \qquad
    \boldsymbol{\tau}_{\min}
    \preceq
    \boldsymbol{\tau}
    \preceq
    \boldsymbol{\tau}_{\max},
\end{equation}
where $M(\mathbf{q}_t)$ is the mass matrix, $\ddot{\mathbf{q}}$ is the joint acceleration, $\mathbf{h}(\mathbf{q}_t,\dot{\mathbf{q}}_t)$ collects Coriolis, centrifugal, and gravity terms, $\boldsymbol{\tau}$ is the joint torque, and $\preceq$ denotes element-wise inequality.
Using the local approximation $\ddot{\mathbf{x}}\approx J_t\ddot{\mathbf{q}}$ and omitting the state-dependent term $\dot{J}_t\dot{\mathbf{q}}_t$ for directional authority estimation, the torque limits induce a task-space acceleration polytope~\cite{chardonnet2009force,skuric2023pycapacity,ferrolho2021residual}:
\begin{equation}
    \mathcal{D}_t
    =
    \left\{
    J_tM(\mathbf{q}_t)^{-1}
    \left(
    \boldsymbol{\tau}
    -
    \mathbf{h}(\mathbf{q}_t,\dot{\mathbf{q}}_t)
    \right)
    \;\middle|\;
    \boldsymbol{\tau}_{\min}
    \preceq
    \boldsymbol{\tau}
    \preceq
    \boldsymbol{\tau}_{\max}
    \right\}.
\end{equation}
This polytope characterizes feasible task-space accelerations under torque limits.
A supporting facet with a smaller distance to the origin indicates weaker acceleration authority along its normal direction.
Let $\{\mathbf{n}_{t,k},\rho_{t,k}\}_{k=1}^{K}$ denote the unit outward normals and support distances of the $K$ supporting facets with the smallest distances to the origin.
We obtain a smooth weak-actuation direction by Boltzmann weighting:
\begin{equation}
    \mathbf{w}_t^{\mathrm{dyn}}
    =
    \sum_{k=1}^{K}
    \alpha_{t,k}\mathbf{n}_{t,k},
    \qquad
    \alpha_{t,k}
    =
    \frac{
    \exp(-\beta \rho_{t,k})
    }
    {
    \sum_{r=1}^{K}
    \exp(-\beta \rho_{t,r})
    },
\end{equation}
where $\beta$ controls the weighting sharpness.

\firstpara{Dynamic degradation loss}
Given the adversarial task-space command $\mathbf{u}_t^{\mathrm{adv}}(P)$, we encourage it to align with the weak-actuation direction:
\begin{equation}
    \ell_{\mathrm{dyn}}
    \left(
        \mathbf{u}_t^{\mathrm{adv}}(P),
        \mathbf{w}_t^{\mathrm{dyn}}
    \right)
    =
    -
    \frac{
    \left(\mathbf{u}_t^{\mathrm{adv}}(P)\right)^{\top}
    \mathbf{w}_t^{\mathrm{dyn}}
    }
    {
    \left\|\mathbf{u}_t^{\mathrm{adv}}(P)\right\|_2
    \left\|\mathbf{w}_t^{\mathrm{dyn}}\right\|_2
    +
    \epsilon
    }.
\end{equation}
Unlike the kinematic loss, no absolute value is used because dynamic actuation authority is direction-dependent.

\subsection{Patch Optimization}
\label{sec:po}

For a rollout sample with observation $I_t$, instruction $L$, and robot state $s_t$, the patched observation induces an adversarial command $\mathbf{c}_t^{\mathrm{adv}}(P)$ and the corresponding task-space command
\begin{equation}
\mathbf{u}_t^{\mathrm{adv}}(P)
=
\Pi(\mathbf{c}_t^{\mathrm{adv}}(P),s_t).
\end{equation}
The weak directions $\mathbf{w}_t^{\mathrm{kin}}$ and $\mathbf{w}_t^{\mathrm{dyn}}$ are computed from the robot state and used as fixed targets during patch optimization.
The execution-degradation surrogate loss is
\begin{equation}
\mathcal{L}_{\mathrm{exec}}(P;I_t,s_t)
=
\lambda
\ell_{\mathrm{kin}}
\left(
\mathbf{u}_t^{\mathrm{adv}}(P),
\mathbf{w}_t^{\mathrm{kin}}
\right)
+
(1-\lambda)
\ell_{\mathrm{dyn}}
\left(
\mathbf{u}_t^{\mathrm{adv}}(P),
\mathbf{w}_t^{\mathrm{dyn}}
\right),
\end{equation}
where $\lambda\in[0,1]$ controls the trade-off between the kinematic and dynamic terms.

Let $\mathcal{I}$ denote the set of observation-state pairs used for patch optimization.
The optimized patch is obtained by
\begin{equation}
P^* = \arg\min_{P\in\mathcal{P}}
\frac{1}{|\mathcal{I}|}
\sum_{(I_t,s_t)\in\mathcal{I}}
\mathcal{L}_{\mathrm{exec}}(P;I_t,s_t).
\end{equation}
During optimization, the VLA policy $\Phi$, instruction, rollout states, downstream controller, and robot model remain fixed; only the patch pixels are updated within $\mathcal{P}$.
For continuous-action policies, $\Pi$ directly extracts the differentiable task-space command.
For tokenized-action policies, the same target direction is enforced through a model-specific prediction loss whose decoded command aligns with the low-executability direction.

\section{Experimental Results}
\label{sec:experiments}

\subsection{Experimental Setup}
\label{sec:experimental_setup}

\firstpara{Implementation}
We implement ExecGap in PyTorch and optimize a $50 \times 50$ RGB patch on $224 \times 224$ observations.
The patch location is randomly sampled during training and fixed to a visible non-occluding location during evaluation.
We use AdamW for 100 epochs with a learning rate of $2 \times 10^{-2}$.
\revise{We select $\lambda=0.15$ using expert trajectories excluded from patch optimization; evaluation rollouts are not used for tuning.}
\revise{For OpenVLA-OFT and $\pi_0$, we compute the surrogate loss using only the first action of each predicted chunk, with low-executability directions estimated at the current robot state.}
All execution-side metrics use translational end-effector displacement, with $\mathbf{u}_t$ and $\mathbf{e}_t$ expressed in the same metric task space.

\firstpara{Benchmarks and evaluation protocol}
We evaluate ExecGap on the four standard 10-task LIBERO suites~\cite{liu2023libero}: Spatial, Object, Goal, and Long.
\revise{For the main comparisons, we sample 10 expert trajectories per task for each victim policy and suite, yielding 100 trajectories to optimize one shared patch for that suite.
Following~\cite{wang2025exploring}, each suite and method in these comparisons is evaluated with 500 online rollouts.}


\firstpara{Victim policies and baselines}
We evaluate three victim VLA policies: OpenVLA~\cite{kim2025openvla}, OpenVLA-OFT~\cite{kim2025fine}, and $\pi_0$~\cite{black2024pi_0}.
For OpenVLA, we compare ExecGap with three visual patch baselines from~\cite{wang2025exploring}: TMA, UADA, and UMA.
\revise{For OpenVLA-OFT and $\pi_0$, we additionally compare against Random Patch (Rand.), TMA, UADA, and UMA.
TMA, UADA, and UMA are directly applied to OpenVLA-OFT.
For $\pi_0$, their action-discrepancy objectives use gradients from the differentiable flow predictor at fixed flow times $t\in\{0,0.2,0.5\}$, without unrolling the full solver.}
Clean denotes unperturbed execution and is not ranked.



\firstpara{Real-robot setting}
We validate ExecGap on a Kinova Jaco2 with its built-in two-finger gripper and a ROKAE xMate ER7 Pro with a Robotiq 2F-85 gripper.
Jaco2 uses one main-view RGB camera, while ROKAE uses both main-view and wrist-mounted RGB cameras.
\revise{The optimized patch is printed and placed at a fixed, visible, non-occluding tabletop location without real-world re-optimization.
For each tested robot--policy pair, we conduct 50 clean trials and 50 attacked trials.}

\firstpara{Evaluation metrics}
We report \textbf{Task Success Rate (TSR)} for task-level disruption, where lower values indicate stronger attacks.
Using the task-space command $\mathbf{u}_t$ and realized execution $\mathbf{e}_t$ defined in Sec.~\ref{sec:problem}, we compute execution-side metrics from translational end-effector displacement over $T$ evaluated execution steps:
\begin{align}
\mathrm{CEDE}
&=
\frac{180}{\pi T}
\sum_{t=1}^{T}
\arccos\!\left(
\operatorname{clip}
\left(
\frac{\mathbf{u}_t^\top\mathbf{e}_t}
{\|\mathbf{u}_t\|_2\|\mathbf{e}_t\|_2+\epsilon},
-1,1
\right)
\right), \\
\mathrm{NCEE}
&=
\frac{1}{T}\sum_{t=1}^{T}
\frac{\bigl|\|\mathbf{u}_t\|_2-\|\mathbf{e}_t\|_2\bigr|}{\|\mathbf{u}_t\|_2+\epsilon},
\qquad
\mathrm{CEG}
=
\frac{1}{T}\sum_{t=1}^{T}
\frac{\|\mathbf{e}_t\|_2}{\|\mathbf{u}_t\|_2+\epsilon}, \\
\mathrm{MRE}
&=
\frac{1}{T}\sum_{t=1}^{T}\|\mathbf{e}_t\|_2,
\qquad
\mathrm{CRE}
=
\sum_{t=1}^{T}\|\mathbf{e}_t\|_2 .
\end{align}
\textbf{Command-Execution Direction Error (CEDE)} measures directional mismatch, while \textbf{Normalized Command-Execution Error (NCEE)} measures normalized command-execution magnitude mismatch.
\textbf{Command-Execution Gain (CEG)} measures realized execution per unit command, while \textbf{Mean Realized Execution (MRE)} and \textbf{Cumulative Realized Execution (CRE)} measure the mean and cumulative amount of realized task-space execution, respectively.
Higher CEDE/NCEE and lower CEG/MRE/CRE indicate stronger execution degradation.

\subsection{Results and Analysis}
\label{sec:results}

\firstpara{Performance on OpenVLA}
Tab.~\ref{tab:main_four_suite_results} reports the main comparison on OpenVLA across four LIBERO suites.
ExecGap reduces TSR to $0.0\%$ on all suites, matching the strongest task-level disruption achieved by prior patch attacks.
Since several baselines already saturate task failure, the key distinction lies in the command-execution metrics.
Across all four suites, ExecGap achieves the largest CEDE and NCEE, indicating stronger directional inconsistency and larger normalized command-execution magnitude mismatch.
It also obtains the lowest CEG on every suite, showing that the predicted commands are less effectively realized by the robot embodiment.
For realized execution, ExecGap achieves the lowest MRE/CRE on Spatial, Object, and Long, and remains close to the best result on Goal.
\revise{These results distinguish ExecGap from prior attacks through stronger command-execution mismatch and weaker command realization, even when TSR is already saturated.}

\begin{table*}[!t]
\centering
\caption{\revise{
Main comparison on OpenVLA across four LIBERO suites.
The table reports task success rate and command-execution metrics.
Best and second-best results among attack methods are marked in \textbf{bold} and \underline{underlined}, respectively.
}}
\label{tab:main_four_suite_results}
\setlength{\tabcolsep}{2.4pt}
\footnotesize
\resizebox{\textwidth}{!}{%
\begin{tabular}{@{}l!{\vrule width 0.5pt}lcccccc!{\vrule width 0.5pt}lcccccc@{}}
\toprule
Attack
& Suite & TSR(\%)$\downarrow$ & CEDE($^\circ$)$\uparrow$ & NCEE$\uparrow$ & CEG$\downarrow$ & MRE$\downarrow$ & CRE$\downarrow$
& Suite & TSR(\%)$\downarrow$ & CEDE($^\circ$)$\uparrow$ & NCEE$\uparrow$ & CEG$\downarrow$ & MRE$\downarrow$ & CRE$\downarrow$ \\
\midrule
Clean
& \multirow{5}{*}{Spatial} & 86.4 & 14.78 & 0.9836 & 0.0165 & 0.0085 & 1.0148
& \multirow{5}{*}{Goal} & 50.2 & 18.30 & 0.9767 & 0.0238 & 0.0045 & 1.5245 \\
TMA
&  & \textbf{0.0} & \underline{21.68} & 0.9824 & 0.0176 & 0.0026 & 0.5620
&  & \textbf{0.0} & \underline{38.13} & 0.9677 & \underline{0.0328} & \textbf{0.0004} & \textbf{0.2095} \\
UADA
&  & \textbf{0.0} & 19.32 & 0.9839 & 0.0161 & 0.0025 & 0.5520
&  & \textbf{0.0} & 34.13 & \underline{0.9692} & 0.0453 & \underline{0.0006} & 0.3016 \\
UMA
&  & \textbf{0.0} & 20.23 & \underline{0.9842} & \underline{0.0158} & \underline{0.0022} & \underline{0.4909}
&  & \textbf{0.0} & 33.53 & 0.9658 & 0.0378 & 0.0007 & 0.3537 \\
Ours
&  & \textbf{0.0} & \textbf{33.72} & \textbf{0.9896} & \textbf{0.0092} & \textbf{0.0020} & \textbf{0.4329}
&  & \textbf{0.0} & \textbf{39.84} & \textbf{0.9930} & \textbf{0.0319} & \textbf{0.0004} & \underline{0.2144} \\
\midrule
Clean
& \multirow{5}{*}{Object} & 76.4 & 14.83 & 0.9742 & 0.0264 & 0.0067 & 1.0760
& \multirow{5}{*}{Long} & 40.8 & 19.21 & 0.9770 & 0.0239 & 0.0043 & 1.5480 \\
TMA
&  & \textbf{0.0} & 25.87 & \underline{0.9807} & \underline{0.0199} & 0.0016 & 0.4567
&  & \underline{1.2} & \underline{35.27} & \underline{0.9696} & \underline{0.0310} & \underline{0.0004} & \underline{0.2037} \\
UADA
&  & \textbf{0.0} & 24.68 & 0.9670 & 0.0338 & \underline{0.0013} & \underline{0.3535}
&  & \textbf{0.0} & 33.62 & 0.9684 & 0.0328 & 0.0006 & 0.2884 \\
UMA
&  & \textbf{0.0} & \underline{27.74} & 0.9780 & 0.0235 & 0.0017 & 0.4790
&  & \textbf{0.0} & 34.00 & 0.9680 & 0.0325 & 0.0006 & 0.3076 \\
Ours
&  & \textbf{0.0} & \textbf{28.54} & \textbf{0.9892} & \textbf{0.0119} & \textbf{0.0010} & \textbf{0.2879}
&  & \textbf{0.0} & \textbf{38.02} & \textbf{0.9854} & \textbf{0.0153} & \textbf{0.0002} & \textbf{0.0784} \\
\bottomrule
\end{tabular}
}
\end{table*}

\begin{figure}[!t]
    \centering
    \includegraphics[width=1\linewidth]{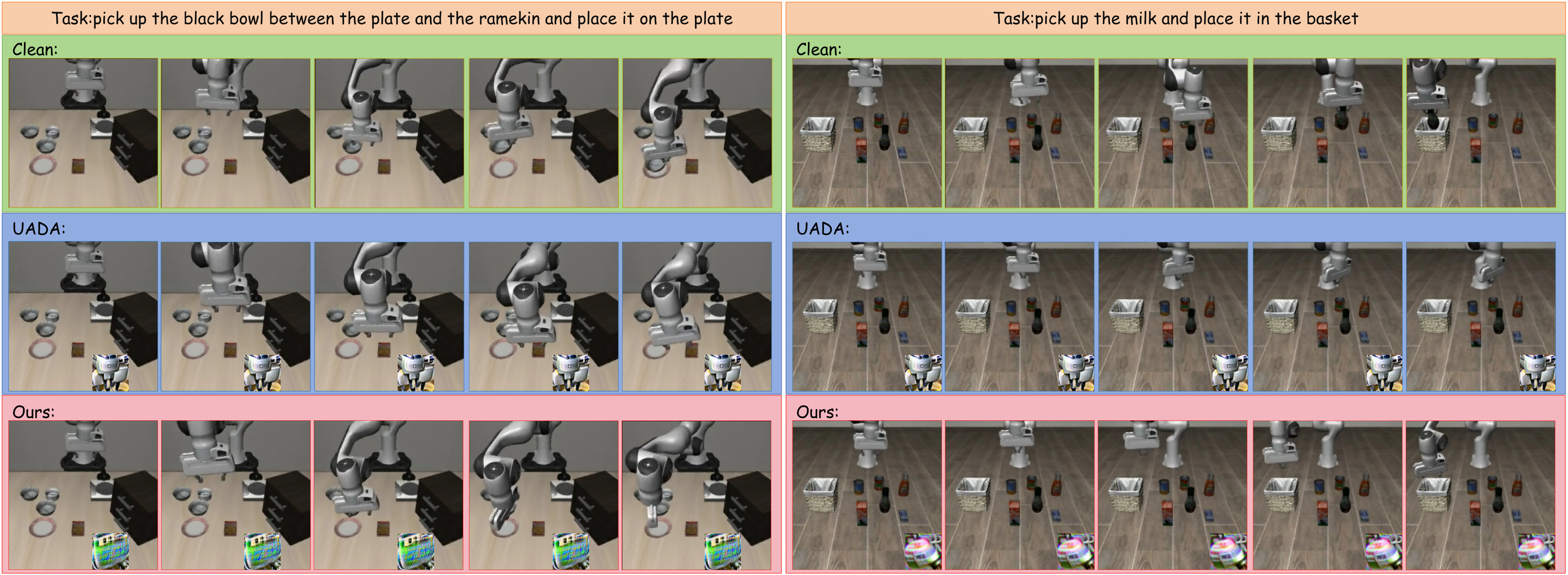}
    \vspace{-3mm}
\caption{
Qualitative rollout snapshots on LIBERO with OpenVLA.
Representative snapshots from OpenVLA-controlled rollouts are shown under clean, UADA, and ExecGap settings.
}
\vspace{-3mm}
    \label{fig:qualitative}
\end{figure}



\firstpara{Qualitative rollout visualization}
Fig.~\ref{fig:qualitative} visualizes representative OpenVLA rollouts on LIBERO under clean, UADA, and ExecGap settings.
Clean execution completes the task, while attacked rollouts fail to do so.
ExecGap often keeps the robot near the task-relevant region but prevents effective final manipulation, illustrating execution degradation through weakened realized execution.

\firstpara{Effectiveness across VLA architectures}
\revise{Tab.~\ref{tab:openvla_oft_pi0_eval} evaluates ExecGap on OpenVLA-OFT and $\pi_0$ across four LIBERO suites.
Compared with Random Patch, ExecGap reduces TSR for both policies on all suites.
Compared with TMA, UADA, and UMA, it yields higher CEDE across all suites for both policies and lower CRE in most settings, indicating stronger command-execution mismatch and weaker realized execution.
Although ExecGap does not always achieve the lowest TSR, these results show that its execution-side effect persists across different VLA architectures and action parameterizations.}

\begin{table*}[!t]
\centering
\caption{\revise{
Comparison on OpenVLA-OFT and $\pi_0$ across four LIBERO suites.
The table reports task success rate and command-execution metrics.
The best result among attack methods for each policy, suite, and metric is marked in \textbf{bold}.
}}
\label{tab:openvla_oft_pi0_eval}
\setlength{\tabcolsep}{2.4pt}
\footnotesize
\resizebox{\textwidth}{!}{%
\begin{tabular}{@{}l!{\vrule width 0.5pt}lccc!{\vrule width 0.5pt}ccc!{\vrule width 0.5pt}lccc!{\vrule width 0.5pt}ccc@{}}
\toprule
\revise{Attack} & \revise{Suite}
& \multicolumn{3}{c!{\vrule width 0.5pt}}{\revise{OpenVLA-OFT}}
& \multicolumn{3}{c!{\vrule width 0.5pt}}{\revise{$\pi_0$}}
& \revise{Suite}
& \multicolumn{3}{c!{\vrule width 0.5pt}}{\revise{OpenVLA-OFT}}
& \multicolumn{3}{c}{\revise{$\pi_0$}} \\
\cmidrule(lr){3-5}\cmidrule(lr){6-8}\cmidrule(lr){10-12}\cmidrule(lr){13-15}
& & \revise{TSR(\%)$\downarrow$} & \revise{CEDE($^\circ$)$\uparrow$} & \revise{CRE$\downarrow$}
& \revise{TSR(\%)$\downarrow$} & \revise{CEDE($^\circ$)$\uparrow$} & \revise{CRE$\downarrow$}
& & \revise{TSR(\%)$\downarrow$} & \revise{CEDE($^\circ$)$\uparrow$} & \revise{CRE$\downarrow$}
& \revise{TSR(\%)$\downarrow$} & \revise{CEDE($^\circ$)$\uparrow$} & \revise{CRE$\downarrow$} \\
\midrule
\revise{Clean} & \multirow{6}{*}{\revise{Spatial}} & \revise{96.2} & \revise{11.54} & \revise{1.0172} & \revise{94.4} & \revise{11.87} & \revise{1.0153} & \multirow{6}{*}{\revise{Goal}} & \revise{98.6} & \revise{19.37} & \revise{1.2181} & \revise{94.4} & \revise{17.93} & \revise{0.8346} \\
\revise{Rand.} &  & \revise{88.6} & \revise{13.65} & \revise{1.1168} & \revise{88.2} & \revise{12.63} & \revise{0.9539} &  & \revise{94.6} & \revise{19.62} & \revise{1.2328} & \revise{90.0} & \revise{17.37} & \revise{0.8682} \\
\revise{TMA} &  & \revise{\textbf{61.6}} & \revise{25.29} & \revise{0.6687} & \revise{68.8} & \revise{27.31} & \revise{0.8153} &  & \revise{72.0} & \revise{21.47} & \revise{0.8474} & \revise{\textbf{64.4}} & \revise{24.58} & \revise{0.7759} \\
\revise{UADA} &  & \revise{66.8} & \revise{25.78} & \revise{0.7351} & \revise{\textbf{65.2}} & \revise{28.33} & \revise{0.7860} &  & \revise{70.0} & \revise{22.93} & \revise{0.7947} & \revise{69.4} & \revise{25.38} & \revise{0.7675} \\
\revise{UMA} &  & \revise{63.2} & \revise{24.28} & \revise{0.6527} & \revise{65.4} & \revise{29.36} & \revise{0.7984} &  & \revise{73.8} & \revise{22.46} & \revise{\textbf{0.7473}} & \revise{75.6} & \revise{24.28} & \revise{0.7234} \\
\revise{Ours} &  & \revise{64.2} & \revise{\textbf{28.75}} & \revise{\textbf{0.6248}} & \revise{68.2} & \revise{\textbf{30.40}} & \revise{\textbf{0.7284}} &  & \revise{\textbf{68.8}} & \revise{\textbf{24.63}} & \revise{0.7486} & \revise{68.2} & \revise{\textbf{29.64}} & \revise{\textbf{0.6889}} \\
\midrule
\revise{Clean} & \multirow{6}{*}{\revise{Object}} & \revise{98.2} & \revise{12.49} & \revise{1.0938} & \revise{99.6} & \revise{10.23} & \revise{1.1483} & \multirow{6}{*}{\revise{Long}} & \revise{92.8} & \revise{18.27} & \revise{1.4287} & \revise{80.2} & \revise{17.63} & \revise{1.9038} \\
\revise{Rand.} &  & \revise{90.6} & \revise{13.72} & \revise{1.0672} & \revise{94.8} & \revise{10.81} & \revise{1.1367} &  & \revise{90.4} & \revise{17.82} & \revise{1.4469} & \revise{74.8} & \revise{18.17} & \revise{1.8297} \\
\revise{TMA} &  & \revise{\textbf{68.2}} & \revise{22.48} & \revise{0.6841} & \revise{67.8} & \revise{25.16} & \revise{0.7468} &  & \revise{58.4} & \revise{25.30} & \revise{0.6139} & \revise{53.2} & \revise{25.54} & \revise{0.3437} \\
\revise{UADA} &  & \revise{68.8} & \revise{24.54} & \revise{0.6645} & \revise{69.6} & \revise{25.13} & \revise{0.7969} &  & \revise{60.8} & \revise{26.35} & \revise{0.6267} & \revise{\textbf{45.0}} & \revise{29.88} & \revise{0.3352} \\
\revise{UMA} &  & \revise{71.2} & \revise{22.13} & \revise{0.7137} & \revise{\textbf{66.4}} & \revise{23.58} & \revise{0.8264} &  & \revise{\textbf{57.6}} & \revise{27.83} & \revise{0.4206} & \revise{47.0} & \revise{28.86} & \revise{0.4730} \\
\revise{Ours} &  & \revise{72.4} & \revise{\textbf{26.72}} & \revise{\textbf{0.5939}} & \revise{66.8} & \revise{\textbf{28.63}} & \revise{\textbf{0.7289}} &  & \revise{60.8} & \revise{\textbf{28.57}} & \revise{\textbf{0.3985}} & \revise{46.6} & \revise{\textbf{34.84}} & \revise{\textbf{0.2507}} \\
\bottomrule
\end{tabular}}
\vspace{-1mm}
\end{table*}

\firstpara{\revise{Effect of commanded motion direction}}
\revise{We compare four choices for the direction of the commanded end-effector displacement $\mathbf{u}_t$ at 400 OpenVLA-OFT states (100 per suite), holding robot state and command magnitude fixed.
Low-executability and high-executability directions correspond to relatively weak and strong estimated motion capability at the current robot state, respectively.
Random uses a randomly sampled task-space direction, while reverse-expert points opposite the end-effector motion in the corresponding expert trajectory.
Tab.~\ref{tab:direction_controls} shows that low-executability directions yield the largest mean CEDE across all four suites, supporting the use of these directions to induce command-execution mismatch.
}


\firstpara{\revise{Generalization to unseen tasks}}
\revise{For each LIBERO suite, we optimize one shared patch using expert trajectories from tasks 0--6 and evaluate it on held-out tasks 7--9 without re-optimization.
Tab.~\ref{tab:heldout_tasks} shows that ExecGap reduces TSR and increases CEDE across all four suites, indicating transfer to tasks excluded from patch optimization within the same suite.}

\begin{table*}[!t]
\centering
\begin{minipage}[t]{0.48\textwidth}
\centering
\captionof{table}{\revise{Effect of commanded motion direction on OpenVLA-OFT.
Mean CEDE ($^\circ$) is reported for each LIBERO suite.
}}
\label{tab:direction_controls}
\setlength{\tabcolsep}{2.5pt}\footnotesize
\renewcommand{\arraystretch}{1.12}
\scalebox{0.89}{%
\begin{tabular}{lcccc}
\toprule
\revise{Suite} & \multicolumn{4}{c}{\revise{Commanded motion direction}} \\
\cmidrule(lr){2-5}
& \revise{Low-exec.} & \revise{Random} & \revise{Reverse-expert} & \revise{High-exec.} \\
\midrule
\revise{Spatial} & \revise{\textbf{24.86}} & \revise{14.63} & \revise{22.74} & \revise{15.31} \\
\revise{Object}  & \revise{\textbf{24.07}} & \revise{15.71} & \revise{21.29} & \revise{16.17} \\
\revise{Goal}    & \revise{\textbf{27.75}} & \revise{17.13} & \revise{26.81} & \revise{17.61} \\
\revise{Long}    & \revise{\textbf{20.28}} & \revise{15.17} & \revise{18.40} & \revise{13.46} \\
\bottomrule
\end{tabular}}
\end{minipage}\hfill
\begin{minipage}[t]{0.48\textwidth}
\centering
\captionof{table}{\revise{Generalization to unseen LIBERO tasks.
Patches are optimized on tasks 0--6 and evaluated on tasks 7--9.
}}
\label{tab:heldout_tasks}
\setlength{\tabcolsep}{3pt}\footnotesize
\scalebox{0.94}{%
\begin{tabular}{lrrrr}
\toprule
\revise{Suite} & \multicolumn{2}{c}{\revise{TSR (\%) $\downarrow$}} & \multicolumn{2}{c}{\revise{CEDE ($^\circ$) $\uparrow$}} \\
\cmidrule(lr){2-3}\cmidrule(lr){4-5}
& \revise{Clean} & \revise{ExecGap} & \revise{Clean} & \revise{ExecGap} \\
\midrule
\revise{Spatial} & \revise{80.0} & \revise{\textbf{33.3}} & \revise{14.96} & \revise{\textbf{22.75}} \\
\revise{Object} & \revise{86.7} & \revise{\textbf{40.0}} & \revise{13.67} & \revise{\textbf{19.97}} \\
\revise{Goal} & \revise{73.3} & \revise{\textbf{33.3}} & \revise{21.70} & \revise{\textbf{28.38}} \\
\revise{Long} & \revise{53.3} & \revise{\textbf{26.7}} & \revise{18.36} & \revise{\textbf{25.44}} \\
\bottomrule
\end{tabular}}
\end{minipage}
\end{table*}

\begin{figure}[!t]
    \centering
    \includegraphics[width=\linewidth]{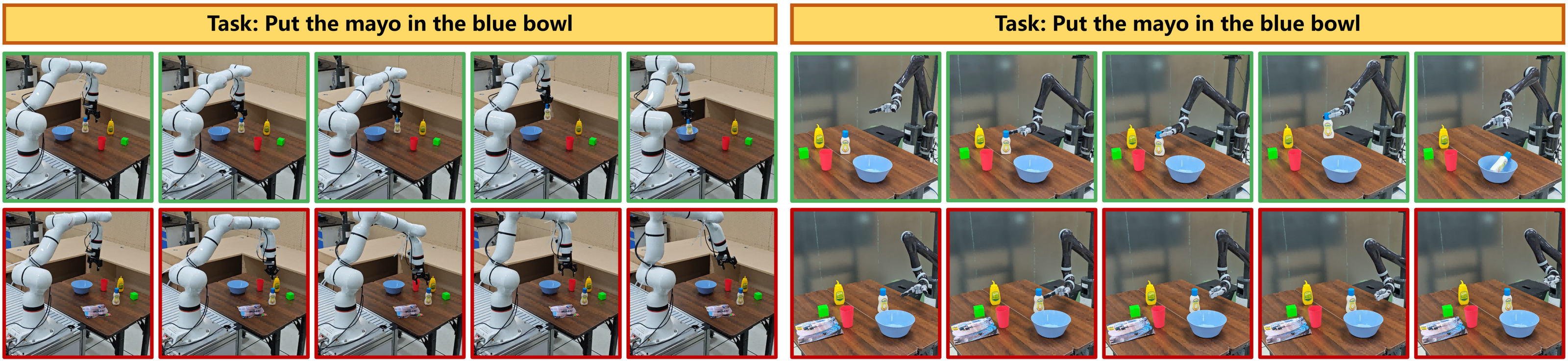}
    \vspace{-3mm}
\caption{Real-robot rollout snapshots. Representative snapshots from OpenVLA-controlled executions on Kinova Jaco2 and ROKAE xMate ER7 Pro are shown under clean and ExecGap settings.}
\vspace{-2mm}
    \label{fig:physical_demo}
\end{figure}

\begin{figure*}[!t]
\centering
\begin{minipage}[t]{0.42\textwidth}
\centering
\includegraphics[width=0.95\linewidth]{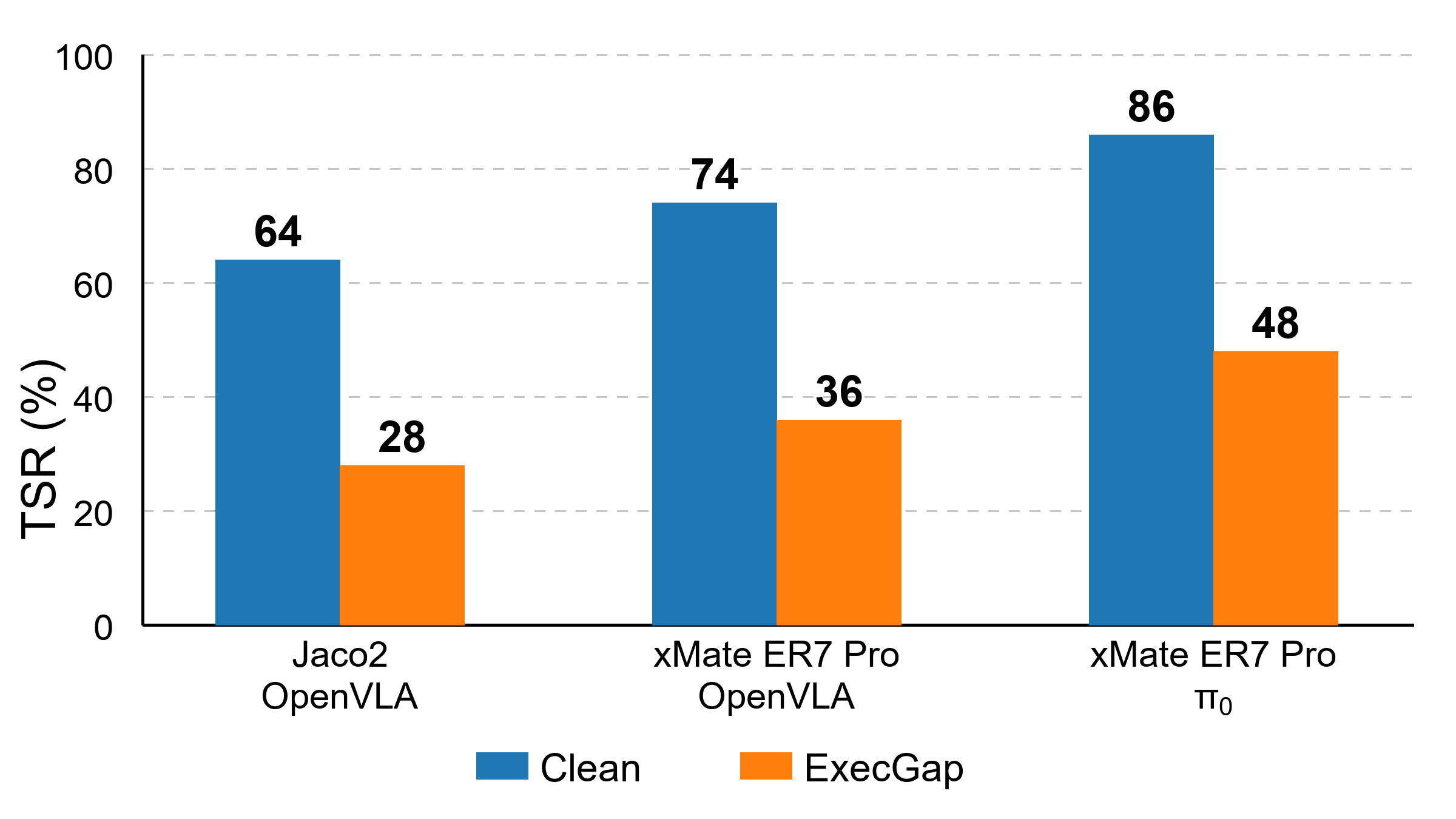}
\captionof{figure}{Real-robot quantitative evaluation. Task success rate is reported under clean and ExecGap settings on the tested robot--policy pairs.}
\label{fig:real_robot_eval}
\end{minipage}\hfill
\begin{minipage}[t]{0.56\textwidth}
\centering
\includegraphics[width=0.78\linewidth]{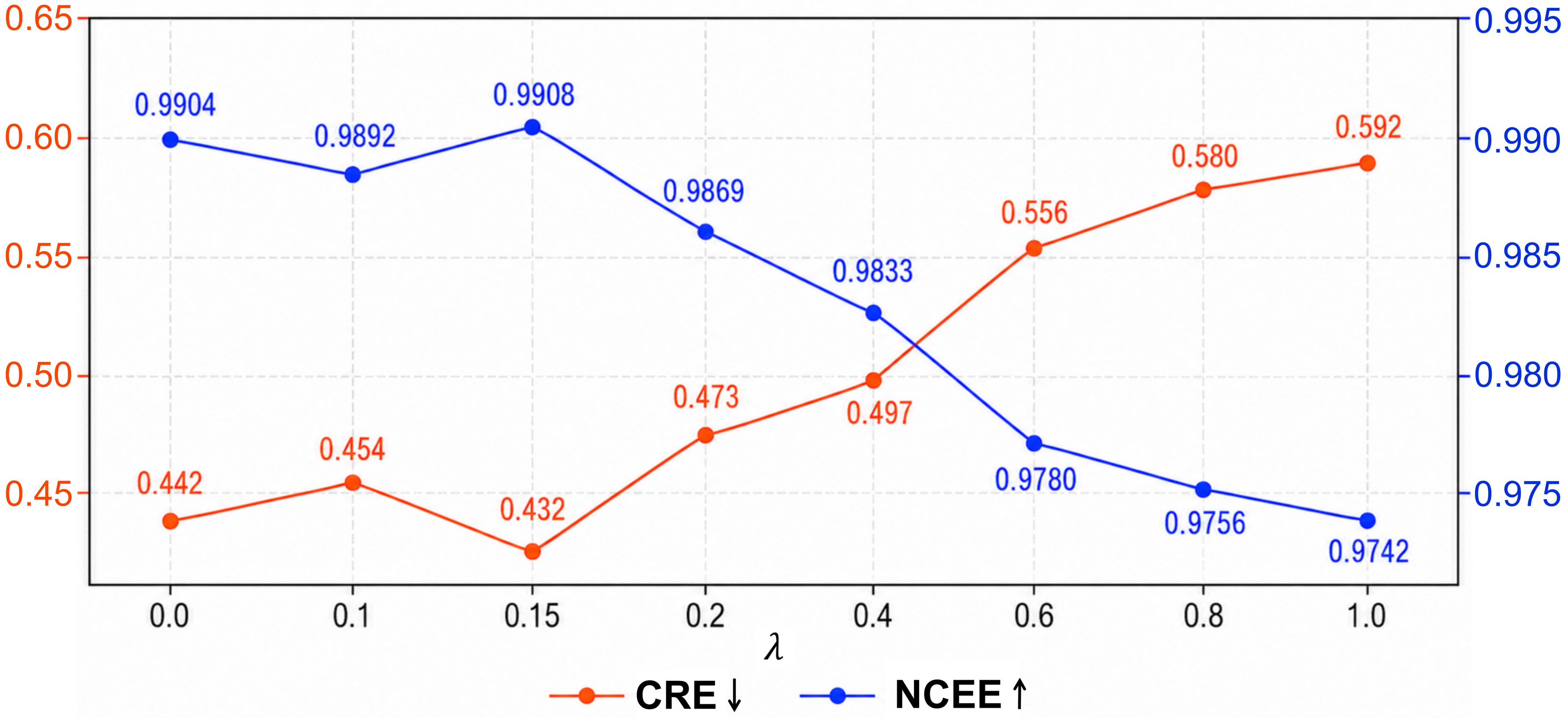}
\captionof{figure}{Component ablation and sensitivity to the kinematic-term weight $\lambda$ in $\mathcal{L}_{\mathrm{exec}}$. CRE and NCEE are reported as $\lambda$ varies.}
\label{fig:lambda_ablation}
\end{minipage}
\end{figure*}

\firstpara{Real-robot validation}
\revise{Fig.~\ref{fig:physical_demo} shows representative real-robot executions under clean and ExecGap settings.
In the attacked examples, the robots approach or manipulate the target but fail to complete the final placement.
Fig.~\ref{fig:real_robot_eval} reports task success over 50 clean and 50 attacked trials for each robot--policy pair.
ExecGap reduces TSR from 64\% to 28\% on Kinova Jaco2 with OpenVLA, from 74\% to 36\% on ROKAE xMate ER7 Pro with OpenVLA, and from 86\% to 48\% on ROKAE xMate ER7 Pro with $\pi_0$.
These results demonstrate the physical feasibility of ExecGap and its ability to disrupt task completion on real robotic platforms.}


\firstpara{Component ablation and weighting sensitivity}
Fig.~\ref{fig:lambda_ablation} evaluates different values of $\lambda$ in $\mathcal{L}_{\mathrm{exec}}$, where $\lambda=0$ and $\lambda=1$ correspond to dynamic-only and kinematic-only variants, respectively.
Introducing dynamic weak-actuation guidance reduces CRE and increases NCEE compared with the kinematic-only variant.
The best performance appears around $\lambda=0.15$, suggesting that dynamic weak actuation complements kinematic inefficiency.

%
%

\section{Conclusion}
\label{sec:conclusion}

This paper has proposed ExecGap, an execution-degradation patch attack for VLA-controlled robots.
ExecGap enlarges command-execution mismatch by biasing predicted task-space commands toward low-executability directions derived from kinematic inefficiency and dynamic weak actuation.
Extensive simulations show that ExecGap induces larger command-execution mismatch and weaker realized execution than existing VLA patch attacks, while real-robot validation confirms that execution degradation can be physically induced.



\section{Limitations}
First, ExecGap requires white-box access to the VLA policy and robot state/model information to estimate low-executability directions.
Second, physical validation is limited to a set of robot platforms and tabletop scenarios.
Finally, this work focuses on exposing execution-degradation attacks rather than studying detection or defense.
\revise{Future work will explore weaker-access settings, broader embodiments and deployments, and execution-aware detection and defense.}

\section*{Acknowledgements}
This work was supported in part by the National Natural Science Foundation of China (62472117, 62402127, 62572400, U2436208,  62372129), the Guangdong Basic and Applied Basic Research Foundation (2025A1515010157, 2024A1515012064), the Science and Technology Projects in Guangzhou
(2025A03J0137, 2024B0101010002), and
 the CCF-NetEase ThunderFire Innovation Research Funding  (CCF-Netease 202514).


\bibliography{references}

\end{document}